\documentclass{article}

\usepackage[final]{nips_2017}

\usepackage[utf8]{inputenc} 
\usepackage[T1]{fontenc}    
\usepackage{hyperref}       
\usepackage{url}            
\usepackage{booktabs}       
\usepackage{amsfonts}       
\usepackage{nicefrac}       
\usepackage{microtype}      
\usepackage{times}  
\usepackage{helvet}  
\usepackage{courier}  
\usepackage{url}  
\usepackage{graphicx}  
\usepackage[utf8]{inputenc} 
\usepackage[T1]{fontenc}    
\usepackage{hyperref}       
\usepackage{url}            
\usepackage{booktabs}       
\usepackage{amsfonts}       
\usepackage{nicefrac}       
\usepackage{microtype}      
\usepackage{soul}

\usepackage[cmex10]{amsmath}
\usepackage{breqn}
\usepackage{array}
\usepackage{fixltx2e}
\usepackage{url}
\usepackage{hyperref}
\usepackage{graphics}
\usepackage{amssymb}
\usepackage{multirow}
\usepackage{tabularx,multicol,multirow}
\usepackage{chngcntr}
\usepackage{color}
\counterwithout*{table}{section}
\counterwithout*{figure}{section}
\usepackage[table,xcdraw]{xcolor}
\usepackage{algorithmicx}
\usepackage{algorithm}
\usepackage{algpseudocode}
\usepackage{varwidth}
\usepackage{setspace}
\algnewcommand\algorithmicinput{\textbf{Input:}}
\algnewcommand\INPUT{\item[\algorithmicinput]}
\algnewcommand\algorithmicoutput{\textbf{Output:}}
\algnewcommand\OUTPUT{\item[\algorithmicoutput]}
\title{RAIL: Risk-Averse Imitation Learning}

\author{Anirban Santara\thanks{Authors contributed equally as a part of their internship at Parallel Computing Lab - Intel Labs, India. }\\
    IIT Kharagpur\\
    \texttt{\small anirban\_santara$@$iitkgp.ac.in}
    \And Abhishek Naik$^{\scriptsize{*}}$ \quad Balaraman Ravindran\\
  IIT Madras\\
  \texttt{\small \{anaik,ravi\}$@$cse.iitm.ac.in} \\
  \AND Dipankar Das \qquad Dheevatsa Mudigere \qquad Sasikanth Avancha \qquad Bharat Kaul\\
  Parallel Computing Lab - Intel Labs, India\\
\texttt{\small{\{dipankar.das,dheevatsa.mudigere,sasikanth.avancha,bharat.kaul\}$@$intel.com}}
}

\begin{document}

\maketitle

\begin{abstract}
Imitation learning algorithms learn viable policies by imitating an expert's behavior when reward signals are not available. Generative Adversarial Imitation Learning (GAIL) is a state-of-the-art algorithm for learning policies when the expert's behavior is available as a fixed set of trajectories. We evaluate in terms of the expert's cost function and observe that the distribution of trajectory-costs is often more heavy-tailed for GAIL-agents than the expert at a number of benchmark continuous-control tasks. Thus, high-cost trajectories, corresponding to tail-end events of catastrophic failure, are more likely to be encountered by the GAIL-agents than the expert. This makes the reliability of GAIL-agents questionable when it comes to deployment in risk-sensitive applications like robotic surgery and autonomous driving. In this work, we aim to minimize the occurrence of tail-end events by minimizing tail risk within the GAIL framework. We quantify tail risk by the Conditional-Value-at-Risk ($CVaR$) of trajectories and develop the Risk-Averse Imitation Learning (RAIL) algorithm. We observe that the policies learned with RAIL show lower tail-end risk than those of vanilla GAIL. Thus the proposed RAIL algorithm appears as a potent alternative to GAIL for improved reliability in risk-sensitive applications. 
\end{abstract}

\begin{figure*}[!t]
\centering
	\includegraphics[width=1.04\linewidth]{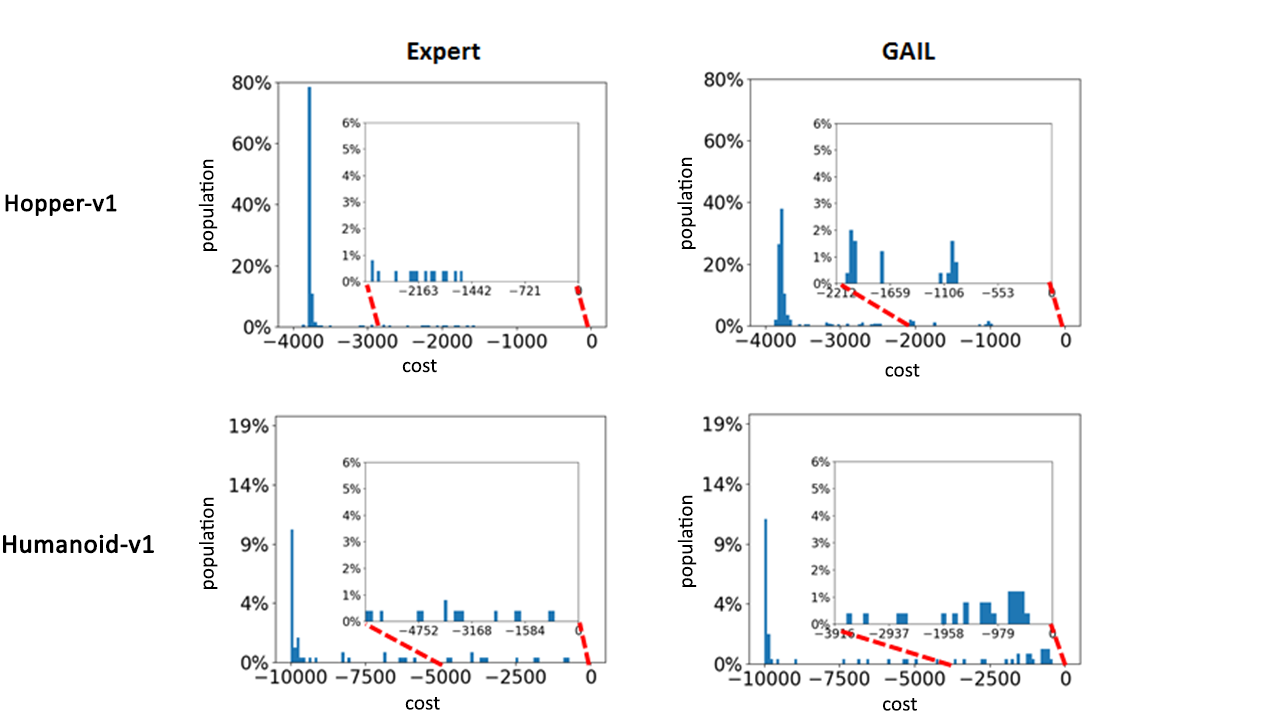}
	\caption{Histograms of the costs of $250$ trajectories generated by the expert and GAIL agents at high-dimensional continuous control tasks, Hopper-v1 and Humanoid-v1, from OpenAI Gym. The inset diagrams show zoomed-in views of the tails of these distributions (the region beyond $2\sigma$ of the mean). We observe that the GAIL agents produce tails heavier than the expert, indicating that GAIL is more prone to generating high-cost trajectories.}
\label{fig:hist}
\end{figure*}%
\section{Introduction}
Reinforcement learning (RL) \citep{Sutton:1998} is used to learn an effective policy of choosing actions in order to achieve a specified goal in an environment. The goal is communicated to the agent through a scalar cost and the agent learns a policy that minimizes the expected total cost incurred over a trajectory. RL algorithms, along with efficient function approximators like deep neural networks, have achieved human-level or beyond human-level performance at many challenging planning tasks like continuous-control \citep{Lillicrap:15,Schulman:15} and game-playing \citep{Silver:2016,Mnih:2015}. In classical RL, the cost function is handcrafted based on heuristic assumptions about the goal and the environment. This is challenging in most real-world applications and also prone to subjectivity induced bias. Imitation learning or Learning from Demonstration (LfD) \citep{Argall:2009,Schaal:1997,Atkeson:1997,Abbeel:2011,Abbeel:2004,Ng:2000} addresses this challenge by providing methods of learning policies through imitation of an expert's behavior without the need of a handcrafted cost function. In this paper we study the reliability of existing imitation learning algorithms when it comes to learning solely from a fixed set of trajectories demonstrated by an expert with no interaction between the agent and the expert during training.

\vspace{-0.25in}
Imitation learning algorithms fall into two broad categories. The first category, known as Behavioral Cloning \citep{Pomerleau:1989,Bojarski:2016,Bojarski:2017}, uses supervised learning to fit a policy function to the state-action pairs from expert-demonstrated trajectories. Despite its simplicity, Behavioral Cloning fails to work well when only a limited amount of data is available. These algorithms assume that observations are i.i.d. and learn to fit single time-step decisions.  Whereas, in sequential decision making problems where predicted actions affect the future observations (e.g. driving), the i.i.d. assumption is violated. As a result, these algorithms suffer from the problem of compounding error due to covariate shift \citep{Ross:2010,Ross:2011}. Approaches to ameliorate the issue of compounding error like SMILe \citep{Ross:2010}, SEARN \citep{daume2009search}, CPI \citep{kakade2002approximately} suffer from instability in practical applications \citep{Ross:2011} while DAGGER \citep{Ross:2011} and AGGREVATE \citep{ross2014reinforcement} require the agent to query the expert during training which is not allowed in our setting of learning from a fixed set of expert demonstrations. Another drawback of Behavioral Cloning is that it does not allow the agent to explore alternate policies for achieving the same objective that might be efficient in some sense other than what the expert cared for.

The second category of algorithms is known as Inverse Reinforcement Learning (IRL) (\citet{Russell:1998,Ng:2000,Abbeel:2011}). It attempts to uncover the underlying reward function that the expert is trying to maximize from a set of expert-demonstrated trajectories. This reward function succinctly encodes the expert's behavior and can be used by an agent to learn a policy through an RL algorithm. The method of learning policies through RL after IRL is known as Apprenticeship Learning (\citet{Abbeel:2004}). IRL algorithms find reward functions that prioritize entire trajectories over others. Unlike behavioral cloning, they do not fit single time-step decisions, and hence they do not suffer from the issue of compounding error. However, IRL algorithms are indirect because they learn a reward function that explains expert behavior but do not tell the learner how to act directly (\citet{Ho:2016}). The job of learning an actionable policy is left to RL algorithms. Moreover, IRL algorithms are computationally expensive and have scalability issues in large environments (\citet{Finn:2016,Levine:2012}).

The recently proposed Generative Adversarial Imitation Learning (GAIL) algorithm \citep{Ho:2016} presents a novel mathematical framework in which the agent learns to act by directly extracting a policy from expert-demonstrated trajectories, as if it were obtained by RL following IRL. The authors show that unlike Behavioral Cloning, this method is not prone to the issue of compounding error and it is also scalable to large environments. Currently, GAIL provides state-of-the-art performance at several benchmark control tasks, including those in Table \ref{table:Hyperparams}.

Risk sensitivity is integral to human learning \citep{nagengast2010risk,niv2012neuralrisk}, and risk-sensitive decision-making problems, in the context of MDPs, have been investigated in various fields, e.g., in finance \citep{ruszczynski2010risk}, operations research \citep{howard1972risk,borkar2002risk}, machine learning \citep{heger1994risk,mihatsch2002risk} and robotics \citep{shalev2016safe,shalev2017formal,abbeel2007application,rajeswaran2016epopt}. \citep{Garcia:2015} give a comprehensive overview of different risk-sensitive RL algorithms. They fall in two broad categories. The first category includes methods that constrain the agent to safe states during exploration while the second modifies the optimality criterion of the agent to embed a term for minimizing risk. Studies on risk-minimization are rather scarce in the imitation learning literature. \citep{Majumdar:2017} take inspiration from studies like \citep{Glimcher:2013,Shen:2014,Hsu:2005} on modeling risk in human decision-making and conservatively approximate the expert's risk preferences by finding an outer approximation of the risk envelope. Much of the literature on imitation learning has been developed with average-case performance at the center, overlooking tail-end events. In this work, we aim to take an inclusive and direct approach to minimizing tail risk of GAIL-learned policies at test time irrespective of the expert's risk preferences. 

In order to evaluate the worst-case risk of deploying GAIL-learned policies, we studied the distributions (see Figure \ref{fig:hist}) of trajectory-costs (according to the expert's cost function) for the GAIL agents and experts at different control tasks (see Table \ref{table:Hyperparams}). We observed that the distributions for GAIL are more heavy-tailed than the expert, where the tail corresponds to occurrences of high trajectory-costs. In order to quantify tail risk, we use Conditional-Value-at-Risk ($CVaR$) \citep{Rockafellar:2000}. $CVaR$ is defined as the expected cost above a given level of confidence and is a popular and coherent tail risk measure. The heavier the tail, the higher the value of $CVaR$. We observe that the value of $CVaR$ is much higher for GAIL than the experts at most of the tasks (see Table \ref{table:Hyperparams}) which again suggests that the GAIL agents encounter high-cost trajectories more often than the experts. Since high trajectory-costs may correspond to events of catastrophic failure, GAIL agents are not reliable in risk-sensitive applications. In this work, we aim to explicitly minimize expected worst-case risk for a given confidence bound (quantified by $CVaR$) along with the GAIL objective, such that the learned policies are more reliable than GAIL, when deployed, while still preserving the average performance of GAIL. \citep{Chow:2014} developed policy gradient and actor-critic algorithms for mean-$CVaR$ optimization for learning policies in the classic RL setting. However these algorithms are not directly applicable in our setting of learning a policy from a set of expert-demonstrated trajectories. We take inspiration from this work and make the following contributions:
\vspace{-3mm}
\begin{enumerate}\itemsep-1mm
\item We formulate the Risk-Averse Imitation Learning (RAIL) algorithm which optimizes $CVaR$ in addition to the original GAIL objective.
\item We evaluate RAIL at a number of benchmark control tasks and demonstrate that it obtains policies with lesser tail risk at test time than GAIL. 
\end{enumerate}
\vspace{-2mm}

The rest of the paper is organized as follows. Section \ref{sec:imitation_learning} builds the mathematical foundation of the paper by introducing essential concepts of imitation learning. Section \ref{sec:RAIL} defines relevant risk-measures and describes the proposed Risk-Averse Imitation Learning algorithm. Section \ref{sec:experimental_setup} specifies our experimental setup and Section \ref{sec:evaluation_metrics} outlines the evaluation metrics. Finally, Section \ref{sec:experimental_results} presents the results of our experiments comparing RAIL with GAIL followed by a discussion of the same and Section \ref{sec:conclusion} concludes the paper with scope of future work.

\vspace{-3mm}
\section{Mathematical Background}
\label{sec:imitation_learning}
\vspace{-2mm}
Let us consider a Markov Decision Process (MDP), $\mathcal{M}=(\mathcal{S}, \mathcal{A}, \mathcal{T}, c, p_0, \gamma)$, where $\mathcal{S}$ denotes the set of all possible states, $\mathcal{A}$ denotes the set of all possible actions that the agent can take, $\mathcal{T}:\mathcal{S}\times\mathcal{A}\times\mathcal{S}\rightarrow[0,1]$ is the state transition function such that, $T(s'|s,a)$ is a probability distribution over next states, $s'\in S$ given current state $s\in S$ and action $a\in A$, $c:\mathcal{S} \times \mathcal{A} \rightarrow \mathbb{R}$ is the cost function which generates a real number as feedback for every state-action pair, $p_0:\mathcal{S}\rightarrow [0,1]$ gives the initial state distribution, and $\gamma$ is a temporal discount factor.

A policy $\pi:\mathcal{S}\times \mathcal{A}\rightarrow [0,1]$ is a function such that $\pi(a|s)$ gives a probability distribution over actions, $a\in \mathcal{A}$ in a given state, $s\in \mathcal{S}$. Let $\xi=(s_0, a_0, s_1, \dots, s_{L_\xi})$ denote a trajectory of length $L_\xi$, obtained by following a policy $\pi$. We define expectation of a function $f(\cdot,\cdot)$ defined on $\mathcal{S}\times\mathcal{A}$ with respect to a policy $\pi$ as follows: 

\vspace{-3mm}

\begin{equation}
\label{eq:PolicyExpectation}
\mathbb{E}_\pi[f(s,a)] \triangleq \mathbb{E}_{\xi\sim \pi} \displaystyle \left[\sum_{t=0}^{L_\xi-1} \gamma^t f(s_t, a_t) \displaystyle \right]
\end{equation}

\subsection{Generative Adversarial Imitation Learning}

Apprenticeship learning or Apprenticeship Learning via Inverse Reinforcement Learning algorithms \citep{Abbeel:2004} first estimate the expert's reward function using IRL and then find the optimal policy for the recovered reward function using RL. Mathematically, this problem can be described as:

\begin{equation}
\label{eq:IL_objective}
RL \circ IRL(\pi_E) = \underset{\pi \in \Pi}{argmin} \max_{c\in \mathcal{C}} \mathbb{E}_{\pi}[c(s,a)] - \mathbb{E}_{\pi_E}[c(s,a)]
-H(\pi)
\end{equation}
\noindent where, $\pi_E$ denotes the expert-policy. $c(\cdot,\cdot)$ denotes the cost function. $\Pi$ and $\mathcal{C}$ denote the hypothesis classes for policy and cost functions. $H(\pi)$ denotes entropy of policy $\pi$. The term $-H(\pi)$ provides causal-entropy regularization \citep{Ziebart:2010,Ziebart:2008} which helps in making the policy optimization algorithm unbiased to factors other than the expected reward.

\citep{Ho:2016} proposed Generative Adversarial Imitation Learning (GAIL) which packs the two step process of $RL \circ IRL_\psi(\pi_E)$ into a single optimization problem with special considerations for scalability in large environments. The name is due to the fact that this objective function can be optimized using the Generative Adversarial Network (GAN) \citep{Goodfellow:2014} framework. The following is objective function of GAIL:
\begin{equation}
\label{eq:GAIL_objective}
\underset{\pi \in \Pi}{argmin} \max_{\mathcal{D}\in(0,1)^{\mathcal{S}\times\mathcal{A}}} \mathbb{E}_\pi[log(\mathcal{D}(s,a))] 
+ \mathbb{E}_{\pi_E}[log(1-\mathcal{D}(s,a))]-H(\pi)
\end{equation}

\noindent Here, the agent's policy, $\pi$, acts as a \emph{generator} of state-action pairs. $\mathcal{D}$ is a discriminative binary classifier of the form $\mathcal{D}:\mathcal{S}\times\mathcal{A}\rightarrow(0,1)$, known as \emph{discriminator}, which given a state-action pair $(s,a)$, predicts the likelihood of it being generated by the generator. A two-player adversarial game is started, wherein the generator tries to generate $(s,a)$ pairs that closely match the expert, while the discriminator tries to correctly classify the $(s,a)$ pairs of the expert and the agent. At convergence, the agent's actions resemble those of the expert in any given state. 

The generator and the discriminator are assigned parameterized models $\pi_\theta$ and $\mathcal{D}_w$ respectively. The training algorithm alternates between a \emph{gradient ascent} step with respect to the discriminator parameters, $w$, and a \emph{policy-gradient descent} step with respect to the generator parameters, $\theta$. Following the example of \citep{Ho:2016} we use multi-layer perceptrons (neural networks with fully-connected layers) \citep{Haykin:1998} to model both the generator and the discriminator.

\section{Risk-Averse Imitation Learning}
\label{sec:RAIL}
In this section, we develop the mathematical formulation of the proposed Risk-Averse Imitation Learning (RAIL) algorithm. We introduce $CVaR$ \citep{Rockafellar:2000} as a measure of tail risk, and apply it in the GAIL-framework to minimize the tail risk of learned policies.

\subsection{Conditional-Value-at-Risk}
\label{subsec:risk}
In the portfolio-risk optimization literature, tail risk is a form of portfolio risk that arises when the possibility that an investment moving more than three standard deviations away from the mean is greater than what is shown by a normal distribution \citep{Investopedia:tailrisk}. Tail risk corresponds to events that have a small probability of occurring. When the distribution of market returns is heavy-tailed, tail risk is high because there is a probability, which may be small, that an investment will move beyond three standard deviations. 

Conditional-Value-at-Risk ($CVaR$) \citep{Rockafellar:2000} is the most conservative measure of tail risk \citep{Dalleh:2011}. Unlike other measures like Variance and Value at Risk ($VaR$), it can be applied when the distribution of returns is not normal. Mathematically, let $Z$ be a random variable. Let $\alpha\in[0,1]$ denote a probability value. The \emph{Value-at-Risk} of $Z$ with respect to confidence level $\alpha$, denoted by $VaR_\alpha(Z)$, is defined as the minimum value $z\in\mathbb{R}$ such that with probability $\alpha$, $Z$ will not exceed $z$.
\begin{equation}
\label{eq:VaR}
VaR_\alpha(Z)=\min(z\ \vert \ P(Z\leq z)\geq \alpha)
\end{equation}
\par \noindent $CVaR_\alpha(Z)$ is defined as the conditional expectation of losses above $VaR_\alpha(Z)$:
\begin{equation}
\label{eq:CVaR}
CVaR_\alpha(Z)= \mathbb{E}\displaystyle\left[ Z\ \vert \ Z\geq VaR_\alpha(Z)\displaystyle\right] = \min_{\nu\in \mathbb{R}} H_\alpha(Z, \nu)
\end{equation}

\noindent where $H_\alpha(Z, \nu)$ is given by:
\vspace{0mm}
\begin{equation}
\label{eq:H_def}
 H_\alpha(Z, \nu) \triangleq  \displaystyle\{\nu+\frac{1}{1-\alpha} \mathbb{E}\left[(Z-\nu)^{+}\right] \displaystyle\};\ (x)^{+}=max(x,0)
\end{equation}

\subsection{RAIL Framework}
\label{subsec:RAILframework}

We use $CVaR$ to quantify the tail risk of the trajectory-cost variable $\mathcal{R}^\pi(\xi|c(\mathcal{D}))$, defined in the context of GAIL as:
\vspace{-3mm}

\begin{equation}
\label{eq:risk}
\mathcal{R}^\pi(\xi|c(\mathcal{D})) = \sum_{t=0}^{L_\xi-1} \gamma^t c(\mathcal{D}(s_t, a_t))
\end{equation}

\noindent where $c(\cdot)$ is order-preserving.

\noindent Next, we formulate the optimization problem to optimize $CVaR$ of $\mathcal{R}^\pi(\xi|c(\mathcal{D}))$ as:
\begin{equation}
\label{eq:minmax_4_CVaR}
\min_\pi \max_c\ CVaR_\alpha(\mathcal{R}^\pi(\xi|c(\mathcal{D})))
=\min_{\pi,\nu} \max_c\  H_\alpha(\mathcal{R}^\pi(\xi|c(\mathcal{D})), \nu) 
\end{equation}

\noindent Integrating this with the GAIL objective of equation \ref{eq:GAIL_objective}, we have the following:
\begin{align}
\label{eq:GAIL_formulation_2_H}
\min_{\pi, \nu} \max_{\mathcal{D}\in(0,1)^{\mathcal{S}\times\mathcal{A}}} \mathcal{J} &= \min_{\pi, \nu} \max_{\mathcal{D}\in(0,1)^{\mathcal{S}\times\mathcal{A}}} \Big\{ -H(\pi) 
+ \mathbb{E}_\pi[log(\mathcal{D}(s,a))] \nonumber \\
&\quad+ \mathbb{E}_{\pi_E}[log(1-\mathcal{D}(s,a))]
+ \lambda_{CVaR}\ H_\alpha(\mathcal{R}^\pi(\xi|c(\mathcal{D})), \nu) \Big\} 
\end{align}

\noindent Note that as $c(\cdot)$ is order-preserving, the maximization with respect to $c$ in equation \ref{eq:minmax_4_CVaR} is equivalent to maximization with respect to $\mathcal{D}$ in equation \ref{eq:GAIL_formulation_2_H}. $\lambda_{CVaR}$ is a constant that controls the amount of weightage given to $CVaR$ optimization relative to the original GAIL objective. Equation \ref{eq:GAIL_formulation_2_H} comprises the objective function of the proposed Risk-Averse Imitation Learning (RAIL) algorithm. Algorithm \ref{algo:RAIL} gives the pseudo-code. Appendix \ref{appendix:gradients} derives the expressions of gradients of the $CVaR$ term $H_\alpha(\mathcal{R}^\pi(\xi|c(\mathcal{D})) \nu)$ with respect to $\pi$, $\mathcal{D}$, and $\nu$. When $\alpha \to 0$, namely the risk-neutral case, $CVaR$ is equal to the mean of all trajectory costs and hence, RAIL $\to$ GAIL. We use Adam algorithm \citep{Kingma:2015} for gradient ascent in the discriminator and Trust Region Policy Optimization (TRPO) \citep{Schulman:15} for policy gradient descent in the generator. The $CVaR$ term $\nu$ is trained by batch gradient descent \citep{Haykin:1998}.

\begin{algorithm}[!h]

\begin{algorithmic}[1]
\caption{Risk-Averse Imitation learning (RAIL)}
\label{algo:RAIL}
\vspace{2mm}
\INPUT \ Expert trajectories $\xi_E\sim\pi_E$, hyper-parameters $\alpha$, $\beta$, $\lambda_{CVaR}$
\OUTPUT Optimized learner's policy $\pi$\\
\vspace{2mm}
Initialization: $\theta \gets \theta_0$, $w\gets w_0$, $\nu\gets \nu_0$, $\lambda\gets \lambda_{CVaR}$
\Repeat
\State Sample trajectories $\xi_i \sim \pi_{\theta_i}$
\begin{spacing}{1.2}
\State Estimate $\hat{H}_\alpha(D^\pi(\xi|c(\mathcal{D})), \nu) = \nu + \frac{1}{1-\alpha} \mathbb{E}_{\xi_i}[\left(D^\pi(\xi|c(\mathcal{D}))-\nu)^{+}\right]$ 
\State \begin{varwidth}[t]{\linewidth}
	Gradient ascent on discriminator parameters using:\par
        \hskip\algorithmicindent $\nabla_{w_{i}}\mathcal{J} = \hat{\mathbb{E}}_{\xi_i}[\nabla_{w_i}\log(\mathcal{D}(s,a))] + \hat{\mathbb{E}}_{\xi_E}[\nabla_{w_i}\log(1-\mathcal{D}(s,a))] $\par
        \hskip\algorithmicindent\hskip11mm $ \;\;\;+\; \lambda_{CVaR} \nabla_{w_i} H_\alpha(\mathcal{R}^\pi(\xi|c(\mathcal{D})), \nu)$ \par 
	\end{varwidth}
\vspace{1mm}
\State \begin{varwidth}[t]{\linewidth}
	KL-constrained natural gradient descent step (TRPO) \\on policy parameters using: \par
    	\hskip\algorithmicindent $\nabla_{\theta_{i}}\mathcal{J} = \mathbb{E}_{(s,a)\sim\xi_i}\left[ \nabla_{\theta_i}log\pi_\theta(a|s)Q(s,a) \right] - \nabla_{\theta_i} H(\pi_\theta) $\par
        \vspace{1mm}\hskip\algorithmicindent\hskip11mm $ \;\;+ \lambda_{CVaR} \nabla_{\theta_i} H_\alpha(\mathcal{R}^\pi(\xi|c(\mathcal{D})), \nu)$ \par
        \vspace{1mm}\hskip\algorithmicindent where 
        $Q(\bar{s},\bar{a}) = \mathbb{E}_{(s,a)\sim\xi_i}[log(\mathcal{D}_{w_{i+1}}(s,a)) | s_0=\bar{s},a_0=\bar{a}]$
    \end{varwidth}
\vspace{1mm}
\State \begin{varwidth}[t]{\linewidth}
	Gradient descent on CVaR parameters: \par
    	\hskip\algorithmicindent $\nabla_{\nu_{i}}\mathcal{J} = \nabla_{\nu_{i}} H_\alpha(\mathcal{R}^\pi(\xi|c(\mathcal{D})), \nu)$ 
	\end{varwidth}
\end{spacing}
\vspace{0mm}
\Until{$i == $ max\_iter}
\end{algorithmic}

\end{algorithm}

\section{Experimental Setup}
\label{sec:experimental_setup}
We compare the tail risk of policies learned by GAIL and RAIL for five continuous control tasks listed in Table \ref{table:Hyperparams}. All these environments, were simulated using MuJoCo Physics Simulator \citep{Todorov:2012}. Each of these environments come packed with a ``true" reward function in OpenAI Gym \citep{Brockman:2016}. \citep{Ho:2016} trained neural network policies using Trust Region Policy Optimization (TRPO) \citep{Schulman:15} on these reward functions to achieve state-of-the-art performance and have made the pre-trained models publicly available for all these environments as a part of their repository \citep{Openai:imitation}. They used these policies to generate the expert trajectories in their work on GAIL \citep{Ho:2016}. For a fair comparison, we use the same policies to generate expert trajectories in our experiments. Table \ref{table:Hyperparams} gives the number of expert trajectories sampled for each environment. These numbers correspond to the best results reported in \citep{Ho:2016}. 

\begin{table}[!b]
\centering
\caption{Hyperparameters for the RAIL experiments on various continuous control tasks from OpenAI Gym. For a fair comparison, the number of training iterations and expert trajectories are same as those used by \citep{Ho:2016}.}
\vspace{3mm}
\label{table:Hyperparams}
\begin{tabular}{|l|c|c|c|}
\hline
\multicolumn{1}{|c|}{\textbf{Task}} & \textbf{\begin{tabular}[c]{@{}c@{}}\#training \\ iterations\end{tabular}} & \textbf{\begin{tabular}[c]{@{}c@{}}\#expert \\ trajectories\end{tabular}} & \multicolumn{1}{l|}{$\lambda_{CVaR}$} \\ \hline
Reacher-v1 & 200 & 18 & 0.25 \\ \hline
HalfCheetah-v1 & 500 & 25 & 0.5 \\ \hline
Hopper-v1 & 500 & 25 & 0.5 \\ \hline
Walker-v1 & 500 & 25 & 0.25 \\ \hline
Humanoid-v1 & 1500 & 240 & 0.75 \\ \hline
\end{tabular}
\end{table}

Again, following \citep{Ho:2016}, we model the generator (policy), discriminator and value function (used for advantage estimation \citep{Sutton:1998} for the generator) with multi-layer perceptrons of the following architecture: \texttt{observationDim - fc\_100 - tanh - fc\_100 - tanh - outDim}, where \texttt{fc\_100} means fully connected layer with 100 nodes, \texttt{tanh} represents the hyperbolic-tangent activation function of the hidden layers, \texttt{observationDim} stands for the dimensionality of the observed feature space, \texttt{outDim} is equal to $1$ for the discriminator and value function networks and equal to the twice of the dimensionality of the action space (for mean and standard deviation of the Gaussian from which the action should be sampled) for the policy network. For example, in case of Humanoid-v1,  \texttt{observationDim} $=376$ and \texttt{outDim} $=34$ in the policy network. The value of the $CVaR$ coefficient $\lambda_{CVaR}$ is set as given by Table \ref{table:Hyperparams} after a coarse hyperparameter search. All other hyperparameters corresponding to the GAIL component of the algorithm are set identical to those used in \citep{Ho:2016} and their repository \citep{Openai:imitation} for all the experiments. The value of $\alpha$ in the $CVaR$ term is set to $0.9$ and its lone parameter, $\nu$, is trained by batch gradient descent with learning rate $0.01$.

\section{Evaluation Metrics}
\label{sec:evaluation_metrics}

In this section we define the metrics we use to evaluate the efficacy of RAIL at reducing the tail risk of GAIL learned policies. Given an agent $A$'s policy $\pi_A$ we roll out $N$ trajectories $T=\{\xi_i\}_{i=1}^N$ from it and estimate $VaR_{\alpha}$ and $CVaR_{\alpha}$ as defined in Section \ref{subsec:risk}. $VaR_{\alpha}$ denotes the value under which the trajectory-cost remains with probability $\alpha$ and $CVaR_{\alpha}$ gives the expected value of cost above $VaR_{\alpha}$. Intuitively, $CVaR_{\alpha}$ gives the average value of cost of the worst cases that have a total probability no more than $(1-\alpha)$. The lower the value of both these metrics, the lower is the tail risk. 

\noindent In order to compare tail risk of an agent with respect to the expert, $E$, we define percentage relative-$VaR_{\alpha}$ as follows:

\begin{equation}
\label{eq:relative-VaR}
VaR_{\alpha}(A|E) = 100\times \frac{VaR_{\alpha}(E) - VaR_{\alpha}(A)}{\vert VaR_{\alpha}(E) \vert} \%
\end{equation}

Similarly, we define percentage relative-$CVaR_{\alpha}$ as:

\begin{equation}
\label{eq:relative-CVaR}
CVaR_{\alpha}(A|E) = 100\times \frac{CVaR_{\alpha}(E) - CVaR_{\alpha}(A)}{\vert CVaR_{\alpha}(E) \vert} \%
\end{equation}

The higher these numbers, the lesser is the tail risk of agent $A$. We define Gain in Reliability (GR) as the difference in percentage relative tail risk between RAIL and GAIL agents.
\begin{align}
\label{eq:GR-VaR}
GR\text{-}VaR &= VaR_{\alpha}(RAIL|E) - VaR_{\alpha}(GAIL|E)\\
\label{eq:GR-CVaR}
GR\text{-}CVaR &= CVaR_{\alpha}(RAIL|E) - CVaR_{\alpha}(GAIL|E)
\end{align}

\begin{figure*}[!t]
\centering
	\includegraphics[width=0.87\linewidth]{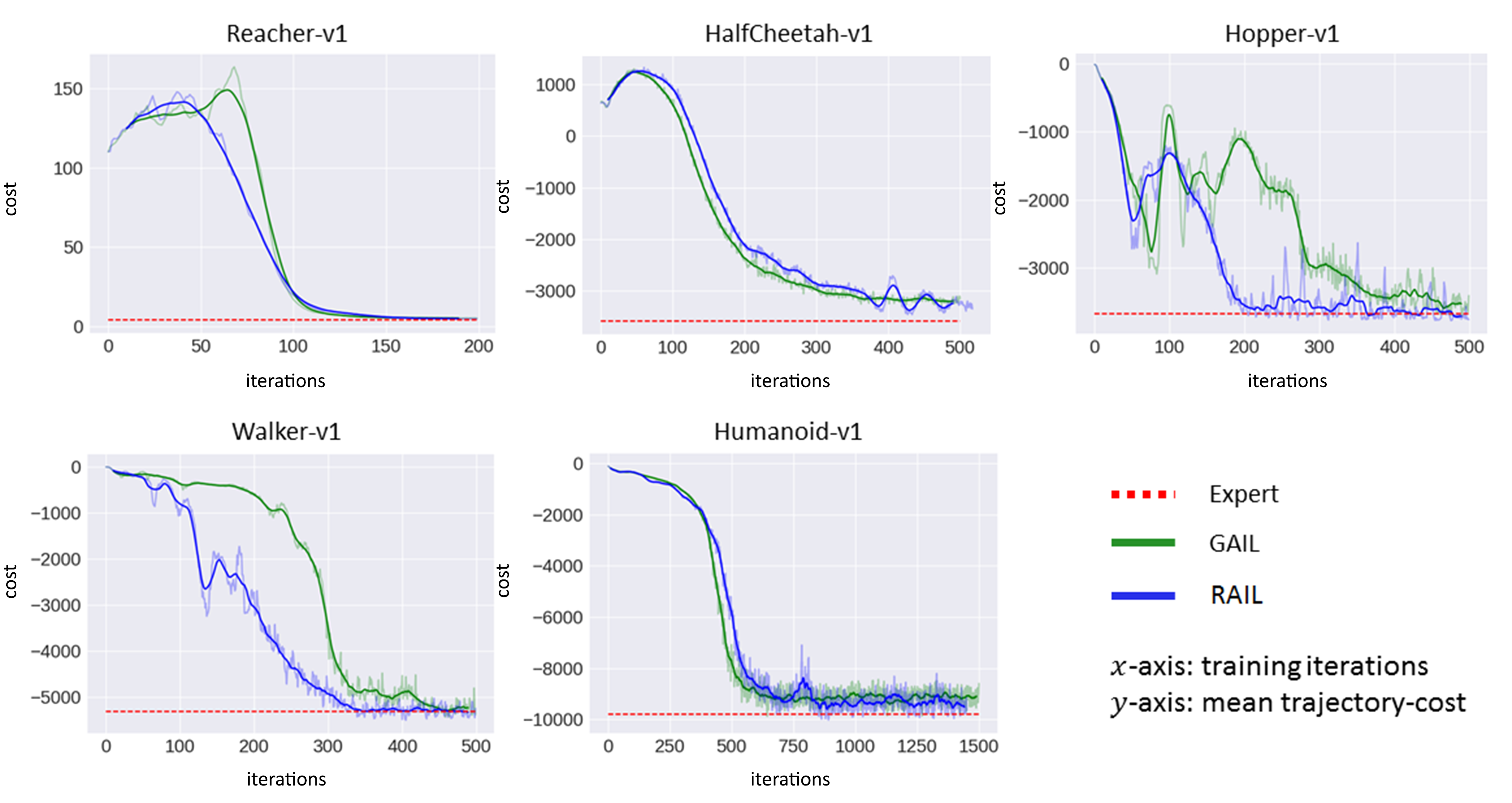}
	\caption{Convergence of mean trajectory-cost during training. The faded curves corresponds to the original value of mean trajectory-cost which varies highly between successive iterations. The data is smoothened with a moving average filter of window size $21$ to demonstrate the prevalent behavior and plotted with solid curves. RAIL converges almost as fast as GAIL at all the five continuous-control tasks, and at times, even faster.}
	\label{fig:mean}
\end{figure*}%

\begin{table*}[!h]
\setlength\tabcolsep{2pt}
\centering
\caption{Comparison of expert, GAIL, and RAIL in terms of the tail risk metrics - $VaR_{0.9}$ and $CVaR_{0.9}$. All the scores are calculated on samples of $50$ trajectories. With smaller values of $VaR$ and $CVaR$, our method outperforms GAIL in all the $5$ continuous control tasks and also outperforms the expert in many cases.}
\vspace{3mm}
\label{table:results}
\begin{tabular}{|l|c|c|c|c|c|c|c|c|}
\hline
\multicolumn{1}{|c|}{\multirow{2}{*}{\textbf{Environment}}} & \multicolumn{2}{c|}{\textbf{Dimensionality}} & \multicolumn{3}{c|}{\textbf{VaR}} & \multicolumn{3}{c|}{\textbf{CVaR}} \\ \cline{2-9} 
\multicolumn{1}{|c|}{} & \textbf{Observation} & \textbf{Action} & \textbf{Expert} & \textbf{GAIL} & \textbf{Ours} & \textbf{Expert} & \textbf{GAIL} & \textbf{Ours} \\ \hline
{Reacher-v1} & 11 & 2 & 5.88 & 9.55 & 7.28 & 6.34 & 13.25 & 9.41 \\ \hline
{Hopper-v1} & 11 & 3 & -3754.71 & -1758.19 & -3745.90 & -2674.65 & -1347.60 & -3727.94 \\ \hline
{HalfCheetah-v1} & 17 & 6 & -3431.59 & -2688.34 & -3150.31 & -3356.67 & -2220.64 & -2945.76 \\ \hline
{Walker-v1} & 17 & 6 & -5402.52 & -5314.05 & -5404.00 & -2310.54 & -3359.29 & -3939.99 \\ \hline
{Humanoid-v1} & 376 & 17 & -9839.79 & -2641.14 & -9252.29 & -4591.43 & -1298.80 & -4640.42 \\ \hline
\end{tabular}
\end{table*}

\begin{table*}[!h]
\centering
\caption{Values of percentage relative tail risk measures and gains in reliability on using RAIL over GAIL for different continuous control tasks.}
\vspace{2mm}
\label{table:derived}
\begin{tabular}{|l|c|c|c|c|c|c|}
\hline
\multicolumn{1}{|c|}{\multirow{2}{*}{\textbf{Environment}}} & \multicolumn{2}{c|}{\textbf{$VaR_{0.9}(A|E) $(\%)}} & \multirow{2}{*}{\textbf{GR-VaR} (\%)} & \multicolumn{2}{c|}{\textbf{$CVaR_{0.9}(A|E)$ (\%)}} & \multirow{2}{*}{\textbf{GR-CVaR} (\%)} \\ \cline{2-3} \cline{5-6}
\multicolumn{1}{|c|}{} & \textbf{GAIL} & \textbf{RAIL} &  & \textbf{GAIL} & \textbf{RAIL} &  \\ \hline 
{Reacher-v1} & -62.41 & -23.81 & \textbf{38.61} & -108.99 & -48.42 & \textbf{60.57} \\ \hline
{Hopper-v1} & -53.17 & -0.23 & \textbf{52.94} & -49.62 & 39.38 & \textbf{89.00} \\ \hline
{HalfCheetah-v1}  & -21.66 & -8.20 & \textbf{13.46} & -33.84 & -12.24 & \textbf{21.60} \\ \hline
{Walker-v1} & -1.64 & 0.03 & \textbf{1.66} & 45.39 & 70.52 & \textbf{25.13} \\ \hline
{Humanoid-v1} & -73.16 & -5.97 & \textbf{67.19} & -71.71 & 1.07 & \textbf{72.78} \\ \hline
\end{tabular}
\end{table*}


\vspace{-2mm}
\section{Experimental Results and Discussion}
\label{sec:experimental_results}
\vspace{-2mm}

In this section, we present and discuss the results of comparison between GAIL and RAIL. The expert's performance is used as a benchmark. Tables \ref{table:results} and \ref{table:derived} present the values of our evaluation metrics for different continuous-control tasks. We set $\alpha=0.9$ for $VaR_{\alpha}$ and $CVaR_{\alpha}$ and estimate all metrics with $N=50$ sampled trajectories (as followed by \citep{Ho:2016}). The following are some interesting observations that we make:
\vspace{-2mm}

\begin{itemize}\itemsep0mm\item RAIL obtains superior performance than GAIL at both tail risk measures -- $VaR_{0.9}$ and $CVaR_{0.9}$, without increasing sample complexity. This shows that RAIL is a superior choice than GAIL for imitation learning in risk-sensitive applications. 

\item The applicability of RAIL is not limited to environments in which the distribution of trajectory-cost is heavy-tailed for GAIL. \citep{Rockafellar:2000} showed that if the distribution of the risk variable $Z$ be normal, $CVaR_{\alpha}(Z) = \mu_Z + a(\alpha)\sigma_Z$, where $a(\alpha)$ is a constant for a given $\alpha$, $\mu_Z$ and $\sigma_Z$ are the mean and standard deviation of $Z$. Thus, in the absence of a heavy tail, minimization of $CVaR_{\alpha}$ of the trajectory cost aids in learning better policies by contributing to the minimization of the mean and standard deviation of trajectory cost. The results on Reacher-v1 corroborate our claims. Although the histogram does not show a heavy tail (Figure \ref{fig:reacher} in Appendix \ref{appendix:analysis}), the mean converges fine (Figure \ref{fig:mean}) and tail risk scores are improved (Table \ref{table:results}) which indicates the distribution of trajectory-costs is more condensed around the mean than GAIL. Thus we can use RAIL instead of GAIL, no matter whether the distribution of trajectory costs is heavy-tailed for GAIL or not.

\item Figure \ref{fig:mean} shows the variation of mean trajectory cost over training iterations for GAIL and RAIL. We observe that RAIL converges almost as fast as GAIL at all the continuous-control tasks in discussion, and at times, even faster. 

\item The success of RAIL in learning a viable policy for Humanoid-v1 suggests that RAIL is scalable to large environments. Scalability is one of the salient features of GAIL. RAIL preserves the scalability of GAIL while showing lower tail risk.
\end{itemize}

\vspace{-1.5mm}
RAIL agents show lesser tail risk than GAIL agents after training has been completed. However it still requires the agent to act in the real world and sample trajectories (line 3 in Algorithm 1) during training. One way to rule out environmental interaction during training is to make the agent act in a simulator while learning from the expert's real-world demonstrations. The setting changes to that of third person imitation learning \citep{stadie2017third}. The RAIL formulation can be easily ported to this framework but we do not evaluate that in this paper.
\vspace{-1mm}

\section{Conclusion}
\label{sec:conclusion}
\vspace{-2mm}
This paper presents the RAIL algorithm which incorporates $CVaR$ optimization within the original GAIL algorithm to minimize tail risk and thus improve the reliability of learned policies. We report significant improvement over GAIL at a number of evaluation metrics on five continuous-control tasks. Thus the proposed algorithm is a viable step in the direction of learning low-risk policies by imitation learning in complex environments, especially in risk-sensitive applications like robotic surgery and autonomous driving. We plan to test RAIL on fielded robotic applications in the future.  


\subsubsection*{Acknowledgments}
\vspace{-2mm}
The authors would like to thank Apoorv Vyas of Intel Labs and Sapana Chaudhary of IIT Madras for helpful discussions. Anirban Santara's travel was supported by Google India under the Google India PhD Fellowship Award.

\bibliography{bibliography}  

\begin{thebibliography}{52}
\providecommand{\natexlab}[1]{#1}
\providecommand{\url}[1]{\texttt{#1}}
\expandafter\ifx\csname urlstyle\endcsname\relax
  \providecommand{\doi}[1]{doi: #1}\else
  \providecommand{\doi}{doi: \begingroup \urlstyle{rm}\Url}\fi

\bibitem[Abbeel and Ng(2004)]{Abbeel:2004}
Pieter Abbeel and Andrew~Y Ng.
\newblock Apprenticeship learning via inverse reinforcement learning.
\newblock In \emph{Proceedings of the twenty-first international conference on
  Machine learning}, page~1. ACM, 2004.

\bibitem[Abbeel and Ng(2011)]{Abbeel:2011}
Pieter Abbeel and Andrew~Y Ng.
\newblock Inverse reinforcement learning.
\newblock In \emph{Encyclopedia of machine learning}, pages 554--558. Springer,
  2011.

\bibitem[Abbeel et~al.(2007)Abbeel, Coates, Quigley, and
  Ng]{abbeel2007application}
Pieter Abbeel, Adam Coates, Morgan Quigley, and Andrew~Y Ng.
\newblock An application of reinforcement learning to aerobatic helicopter
  flight.
\newblock In \emph{Advances in neural information processing systems}, pages
  1--8, 2007.

\bibitem[Argall et~al.(2009)Argall, Chernova, Veloso, and
  Browning]{Argall:2009}
Brenna~D. Argall, Sonia Chernova, Manuela Veloso, and Brett Browning.
\newblock A survey of robot learning from demonstration.
\newblock \emph{Robotics and Autonomous Systems}, 57\penalty0 (5):\penalty0 469
  -- 483, 2009.
\newblock ISSN 0921-8890.
\newblock \doi{http://dx.doi.org/10.1016/j.robot.2008.10.024}.
\newblock URL
  \url{http://www.sciencedirect.com/science/article/pii/S0921889008001772}.

\bibitem[Atkeson and Schaal(1997)]{Atkeson:1997}
Christopher~G Atkeson and Stefan Schaal.
\newblock Robot learning from demonstration.
\newblock In \emph{ICML}, volume~97, pages 12--20, 1997.

\bibitem[Bojarski et~al.(2016)Bojarski, Del~Testa, Dworakowski, Firner, Flepp,
  Goyal, Jackel, Monfort, Muller, Zhang, et~al.]{Bojarski:2016}
Mariusz Bojarski, Davide Del~Testa, Daniel Dworakowski, Bernhard Firner, Beat
  Flepp, Prasoon Goyal, Lawrence~D Jackel, Mathew Monfort, Urs Muller, Jiakai
  Zhang, et~al.
\newblock End to end learning for self-driving cars.
\newblock \emph{arXiv preprint arXiv:1604.07316}, 2016.

\bibitem[Bojarski et~al.(2017)Bojarski, Yeres, Choromanska, Choromanski,
  Firner, Jackel, and Muller]{Bojarski:2017}
Mariusz Bojarski, Philip Yeres, Anna Choromanska, Krzysztof Choromanski,
  Bernhard Firner, Lawrence Jackel, and Urs Muller.
\newblock Explaining how a deep neural network trained with end-to-end learning
  steers a car.
\newblock \emph{arXiv preprint arXiv:1704.07911}, 2017.

\bibitem[Borkar(2002)]{borkar2002risk}
Vivek~S Borkar.
\newblock Q-learning for risk-sensitive control.
\newblock \emph{Mathematics of operations research}, 27\penalty0 (2):\penalty0
  294--311, 2002.

\bibitem[Brockman et~al.(2016)Brockman, Cheung, Pettersson, Schneider,
  Schulman, Tang, and Zaremba]{Brockman:2016}
Greg Brockman, Vicki Cheung, Ludwig Pettersson, Jonas Schneider, John Schulman,
  Jie Tang, and Wojciech Zaremba.
\newblock Openai gym.
\newblock \emph{arXiv preprint arXiv:1606.01540}, 2016.

\bibitem[Chow and Ghavamzadeh(2014)]{Chow:2014}
Yinlam Chow and Mohammad Ghavamzadeh.
\newblock Algorithms for cvar optimization in mdps.
\newblock In \emph{Advances in neural information processing systems}, pages
  3509--3517, 2014.

\bibitem[Dalleh(2011)]{Dalleh:2011}
Nivine Dalleh.
\newblock \emph{Why is CVaR superior to VaR?(c2009)}.
\newblock PhD thesis, 2011.

\bibitem[Daum{\'e} et~al.(2009)Daum{\'e}, Langford, and Marcu]{daume2009search}
Hal Daum{\'e}, John Langford, and Daniel Marcu.
\newblock Search-based structured prediction.
\newblock \emph{Machine learning}, 75\penalty0 (3):\penalty0 297--325, 2009.

\bibitem[Diederik~Kingma(2015)]{Kingma:2015}
Jimmy~Ba Diederik~Kingma.
\newblock Adam: A method for stochastic optimization.
\newblock \emph{arXiv:1310.5107 [cs.CV]}, 2015.

\bibitem[Finn et~al.(2016)Finn, Levine, and Abbeel]{Finn:2016}
Chelsea Finn, Sergey Levine, and Pieter Abbeel.
\newblock Guided cost learning: Deep inverse optimal control via policy
  optimization.
\newblock In \emph{International Conference on Machine Learning}, pages 49--58,
  2016.

\bibitem[Garc{\i}a and Fern{\'a}ndez(2015)]{Garcia:2015}
Javier Garc{\i}a and Fernando Fern{\'a}ndez.
\newblock A comprehensive survey on safe reinforcement learning.
\newblock \emph{Journal of Machine Learning Research}, 16\penalty0
  (1):\penalty0 1437--1480, 2015.

\bibitem[Glimcher and Fehr(2013)]{Glimcher:2013}
Paul~W Glimcher and Ernst Fehr.
\newblock \emph{Neuroeconomics: Decision making and the brain}.
\newblock Academic Press, 2013.

\bibitem[Goodfellow et~al.(2014)Goodfellow, Pouget-Abadie, Mirza, Xu,
  Warde-Farley, Ozair, Courville, and Bengio]{Goodfellow:2014}
Ian Goodfellow, Jean Pouget-Abadie, Mehdi Mirza, Bing Xu, David Warde-Farley,
  Sherjil Ozair, Aaron Courville, and Yoshua Bengio.
\newblock Generative adversarial nets.
\newblock In \emph{Advances in neural information processing systems}, pages
  2672--2680, 2014.

\bibitem[Haykin(1998)]{Haykin:1998}
Simon Haykin.
\newblock \emph{Neural Networks: A Comprehensive Foundation}.
\newblock Prentice Hall PTR, Upper Saddle River, NJ, USA, 2nd edition, 1998.
\newblock ISBN 0132733501.

\bibitem[Heger(1994)]{heger1994risk}
Matthias Heger.
\newblock Consideration of risk in reinforcement learning.
\newblock In \emph{Proceedings of the Eleventh International Conference on
  Machine Learning}, pages 105--111, 1994.

\bibitem[Ho and Ermon(2016)]{Ho:2016}
Jonathan Ho and Stefano Ermon.
\newblock Generative adversarial imitation learning.
\newblock In \emph{Advances in Neural Information Processing Systems}, pages
  4565--4573, 2016.

\bibitem[Howard and Matheson(1972)]{howard1972risk}
Ronald~A Howard and James~E Matheson.
\newblock Risk-sensitive markov decision processes.
\newblock \emph{Management science}, 18\penalty0 (7):\penalty0 356--369, 1972.

\bibitem[Hsu et~al.(2005)Hsu, Bhatt, Adolphs, Tranel, and Camerer]{Hsu:2005}
Ming Hsu, Meghana Bhatt, Ralph Adolphs, Daniel Tranel, and Colin~F Camerer.
\newblock Neural systems responding to degrees of uncertainty in human
  decision-making.
\newblock \emph{Science}, 310\penalty0 (5754):\penalty0 1680--1683, 2005.

\bibitem[Investopedia(2017)]{Investopedia:tailrisk}
Investopedia.
\newblock Definition of tail risk.
\newblock \url{http://www.investopedia.com/terms/t/tailrisk.asp}, 2017.
\newblock Accessed: 2017-09-11.

\bibitem[Kakade and Langford(2002)]{kakade2002approximately}
Sham Kakade and John Langford.
\newblock Approximately optimal approximate reinforcement learning.
\newblock In \emph{ICML}, volume~2, pages 267--274, 2002.

\bibitem[Levine and Koltun(2012)]{Levine:2012}
Sergey Levine and Vladlen Koltun.
\newblock Continuous inverse optimal control with locally optimal examples.
\newblock \emph{arXiv preprint arXiv:1206.4617}, 2012.

\bibitem[Lillicrap et~al.(2015)Lillicrap, Hunt, Pritzel, Heess, Erez, Tassa,
  Silver, and Wierstra]{Lillicrap:15}
Timothy~P. Lillicrap, Jonathan~J. Hunt, Alexander Pritzel, Nicolas Heess, Tom
  Erez, Yuval Tassa, David Silver, and Daan Wierstra.
\newblock Continuous control with deep reinforcement learning.
\newblock \emph{CoRR}, abs/1509.02971, 2015.
\newblock URL \url{http://arxiv.org/abs/1509.02971}.

\bibitem[Majumdar et~al.(2017)Majumdar, Singh, Mandlekar, and
  Pavone]{Majumdar:2017}
Anirudha Majumdar, Sumeet Singh, Ajay Mandlekar, and Marco Pavone.
\newblock Risk-sensitive inverse reinforcement learning via coherent risk
  models.
\newblock 2017.

\bibitem[Mihatsch and Neuneier(2002)]{mihatsch2002risk}
Oliver Mihatsch and Ralph Neuneier.
\newblock Risk-sensitive reinforcement learning.
\newblock \emph{Machine learning}, 49\penalty0 (2-3):\penalty0 267--290, 2002.

\bibitem[Mnih et~al.(2015)Mnih, Kavukcuoglu, Silver, Rusu, Veness, Bellemare,
  Graves, Riedmiller, Fidjeland, Ostrovski, et~al.]{Mnih:2015}
Volodymyr Mnih, Koray Kavukcuoglu, David Silver, Andrei~A Rusu, Joel Veness,
  Marc~G Bellemare, Alex Graves, Martin Riedmiller, Andreas~K Fidjeland, Georg
  Ostrovski, et~al.
\newblock Human-level control through deep reinforcement learning.
\newblock \emph{Nature}, 518\penalty0 (7540):\penalty0 529--533, 2015.

\bibitem[Nagengast et~al.(2010)Nagengast, Braun, and
  Wolpert]{nagengast2010risk}
Arne~J Nagengast, Daniel~A Braun, and Daniel~M Wolpert.
\newblock Risk-sensitive optimal feedback control accounts for sensorimotor
  behavior under uncertainty.
\newblock \emph{PLoS computational biology}, 6\penalty0 (7):\penalty0 e1000857,
  2010.

\bibitem[Ng et~al.(2000)Ng, Russell, et~al.]{Ng:2000}
Andrew~Y Ng, Stuart~J Russell, et~al.
\newblock Algorithms for inverse reinforcement learning.
\newblock In \emph{Icml}, pages 663--670, 2000.

\bibitem[Niv et~al.(2012)Niv, Edlund, Dayan, and O'Doherty]{niv2012neuralrisk}
Yael Niv, Jeffrey~A Edlund, Peter Dayan, and John~P O'Doherty.
\newblock Neural prediction errors reveal a risk-sensitive
  reinforcement-learning process in the human brain.
\newblock \emph{Journal of Neuroscience}, 32\penalty0 (2):\penalty0 551--562,
  2012.

\bibitem[OpenAI-GAIL(2017)]{Openai:imitation}
OpenAI-GAIL.
\newblock Imitation learning github repository.
\newblock \url{https://github.com/openai/imitation.git}, 2017.
\newblock Accessed: 2017-06-27.

\bibitem[Pomerleau(1989)]{Pomerleau:1989}
Dean~A Pomerleau.
\newblock Alvinn: An autonomous land vehicle in a neural network.
\newblock In \emph{Advances in neural information processing systems}, pages
  305--313, 1989.

\bibitem[Rajeswaran et~al.(2016)Rajeswaran, Ghotra, Levine, and
  Ravindran]{rajeswaran2016epopt}
Aravind Rajeswaran, Sarvjeet Ghotra, Sergey Levine, and Balaraman Ravindran.
\newblock Epopt: Learning robust neural network policies using model ensembles.
\newblock \emph{5th International Conference on Learning Representations},
  2016.

\bibitem[Rockafellar and Uryasev(2000)]{Rockafellar:2000}
R~Tyrrell Rockafellar and Stanislav Uryasev.
\newblock Optimization of conditional value-at-risk.
\newblock \emph{Journal of risk}, 2:\penalty0 21--42, 2000.

\bibitem[Ross and Bagnell(2010)]{Ross:2010}
St{\'e}phane Ross and Drew Bagnell.
\newblock Efficient reductions for imitation learning.
\newblock In \emph{Proceedings of the thirteenth international conference on
  artificial intelligence and statistics}, pages 661--668, 2010.

\bibitem[Ross and Bagnell(2014)]{ross2014reinforcement}
Stephane Ross and J~Andrew Bagnell.
\newblock Reinforcement and imitation learning via interactive no-regret
  learning.
\newblock \emph{arXiv preprint arXiv:1406.5979}, 2014.

\bibitem[Ross et~al.(2011)Ross, Gordon, and Bagnell]{Ross:2011}
St{\'e}phane Ross, Geoffrey~J Gordon, and Drew Bagnell.
\newblock A reduction of imitation learning and structured prediction to
  no-regret online learning.
\newblock In \emph{International Conference on Artificial Intelligence and
  Statistics}, pages 627--635, 2011.

\bibitem[Russell(1998)]{Russell:1998}
Stuart Russell.
\newblock Learning agents for uncertain environments.
\newblock In \emph{Proceedings of the eleventh annual conference on
  Computational learning theory}, pages 101--103. ACM, 1998.

\bibitem[Ruszczy{\'n}ski(2010)]{ruszczynski2010risk}
Andrzej Ruszczy{\'n}ski.
\newblock Risk-averse dynamic programming for markov decision processes.
\newblock \emph{Mathematical programming}, 125\penalty0 (2):\penalty0 235--261,
  2010.

\bibitem[Schaal(1997)]{Schaal:1997}
Stefan Schaal.
\newblock Learning from demonstration.
\newblock In \emph{Advances in neural information processing systems}, pages
  1040--1046, 1997.

\bibitem[Schulman et~al.(2015)Schulman, Levine, Moritz, Jordan, and
  Abbeel]{Schulman:15}
John Schulman, Sergey Levine, Philipp Moritz, Michael~I. Jordan, and Pieter
  Abbeel.
\newblock Trust region policy optimization.
\newblock \emph{CoRR}, abs/1502.05477, 2015.
\newblock URL \url{http://arxiv.org/abs/1502.05477}.

\bibitem[Shalev-Shwartz et~al.(2016)Shalev-Shwartz, Shammah, and
  Shashua]{shalev2016safe}
Shai Shalev-Shwartz, Shaked Shammah, and Amnon Shashua.
\newblock Safe, multi-agent, reinforcement learning for autonomous driving.
\newblock \emph{arXiv preprint arXiv:1610.03295}, 2016.

\bibitem[Shalev-Shwartz et~al.(2017)Shalev-Shwartz, Shammah, and
  Shashua]{shalev2017formal}
Shai Shalev-Shwartz, Shaked Shammah, and Amnon Shashua.
\newblock On a formal model of safe and scalable self-driving cars.
\newblock \emph{arXiv preprint arXiv:1708.06374}, 2017.

\bibitem[Shen et~al.(2014)Shen, Tobia, Sommer, and Obermayer]{Shen:2014}
Yun Shen, Michael~J Tobia, Tobias Sommer, and Klaus Obermayer.
\newblock Risk-sensitive reinforcement learning.
\newblock \emph{Neural computation}, 26\penalty0 (7):\penalty0 1298--1328,
  2014.

\bibitem[Silver et~al.(2016)Silver, Huang, Maddison, Guez, Sifre, Van
  Den~Driessche, Schrittwieser, Antonoglou, Panneershelvam, Lanctot,
  et~al.]{Silver:2016}
David Silver, Aja Huang, Chris~J Maddison, Arthur Guez, Laurent Sifre, George
  Van Den~Driessche, Julian Schrittwieser, Ioannis Antonoglou, Veda
  Panneershelvam, Marc Lanctot, et~al.
\newblock Mastering the game of go with deep neural networks and tree search.
\newblock \emph{Nature}, 529\penalty0 (7587):\penalty0 484--489, 2016.

\bibitem[Stadie et~al.(2017)Stadie, Abbeel, and Sutskever]{stadie2017third}
Bradly~C Stadie, Pieter Abbeel, and Ilya Sutskever.
\newblock Third-person imitation learning.
\newblock \emph{arXiv preprint arXiv:1703.01703}, 2017.

\bibitem[Sutton and Barto(1998)]{Sutton:1998}
R.S. Sutton and A.G. Barto.
\newblock \emph{Reinforcement Learning: An Introduction}.
\newblock A Bradford book. Bradford Book, 1998.
\newblock ISBN 9780262193986.
\newblock URL \url{https://books.google.co.in/books?id=CAFR6IBF4xYC}.

\bibitem[Todorov et~al.(2012)Todorov, Erez, and Tassa]{Todorov:2012}
Emanuel Todorov, Tom Erez, and Yuval Tassa.
\newblock Mujoco: A physics engine for model-based control.
\newblock In \emph{Intelligent Robots and Systems (IROS), 2012 IEEE/RSJ
  International Conference on}, pages 5026--5033. IEEE, 2012.

\bibitem[Ziebart(2010)]{Ziebart:2010}
Brian~D Ziebart.
\newblock \emph{Modeling Purposeful Adaptive Behavior with the Principle of
  Maximum Causal Entropy}.
\newblock PhD thesis, Carnegie Mellon University, 2010.

\bibitem[Ziebart et~al.(2008)Ziebart, Maas, Bagnell, and Dey]{Ziebart:2008}
Brian~D Ziebart, Andrew~L Maas, J~Andrew Bagnell, and Anind~K Dey.
\newblock Maximum entropy inverse reinforcement learning.
\newblock In \emph{AAAI}, volume~8, pages 1433--1438. Chicago, IL, USA, 2008.

\end{thebibliography}
\bibliographystyle{plainnat}


\newpage

\section*{{Appendix}}
\label{sec:appendix}
\renewcommand{\thesubsection}{\Alph{subsection}}

\counterwithin*{equation}{subsection}
\renewcommand{\theequation}{\thesubsection.\arabic{equation}}


\subsection{Calculation of Gradients of the CVaR term}
\label{appendix:gradients}

In this section we derive expressions of gradients of the CVaR term in equation \ref{eq:GAIL_formulation_2_H} w.r.t. $\pi$, $\mathcal{D}$, and $\nu$. \\Let us denote $H_\alpha(D^\pi(\xi|c(\mathcal{D})), \nu)$ by $\mathcal{L}_{CVaR}$. Our derivations are inspired by those shown by \citet{Chow:2014}.

\begin{itemize}

\item \textbf{Gradient of $\mathcal{L}_{CVaR}$ w.r.t. $\mathcal{D}$:}

\begin{align}
\label{eq:grad_L_D_1}
\nabla_{\mathcal{D}}\ \mathcal{L}_{CVaR} & = \nabla_{\mathcal{D}}\ \displaystyle\left[ \nu + \frac{1}{1-\alpha} \mathbb{E}_{\xi\sim\pi}\left[ (D^\pi(\xi|c(\mathcal{D}))-\nu)^{+}\right] \displaystyle\right] \nonumber \\
& = \frac{1}{1-\alpha}\mathbb{E}_{\xi\sim\pi}\left[ \nabla_{\mathcal{D}}\ D^\pi(\xi|c(\mathcal{D}))\mathbf{1}(D^\pi(\xi|c(\mathcal{D}))\geq\nu)\right]
\end{align}
\noindent where $\mathbf{1}(\cdot)$ denotes the \emph{indicator function}. Now,

\begin{align}
\label{eq:grad_L_D_2}
\nabla_{\mathcal{D}}\ D^\pi(\xi|c(\mathcal{D})) &= \nabla_{c}\ D^\pi(\xi|c(\mathcal{D}))\ \nabla_{\mathcal{D}}\ c(\mathcal{D})\\
\label{eq:grad_L_D_3}
\nabla_{c}\ D^\pi(\xi|c(\mathcal{D})) &=  \nabla_{c}\ \sum_{t=0}^{L_\xi-1} \gamma^t c(s_t, a_t) \nonumber \\
&=  \sum_{t=0}^{L_\xi-1} \gamma^t \nonumber \\
&=  \frac{1-\gamma^{L_\xi}}{1-\gamma}
\end{align}

Substituting equation \ref{eq:grad_L_D_3} in \ref{eq:grad_L_D_2} and then \ref{eq:grad_L_D_2} in \ref{eq:grad_L_D_1}, we have the following:

\begin{equation}
\nabla_{\mathcal{D}}\ \mathcal{L}_{CVaR} = \frac{1}{1-\alpha}\mathbb{E}_{\xi\sim\pi}\left[ \frac{1-\gamma^{L_\xi}}{1-\gamma}\ \mathbf{1}(D^\pi(\xi|c(\mathcal{D}))\geq\nu) 
\nabla_{\mathcal{D}}\ c(\mathcal{D})\ \right]
\end{equation}
\vspace{4mm}

\item \textbf{Gradient of $\mathcal{L}_{CVaR}$ w.r.t. $\pi$:}
\begin{eqnarray}
\label{eq:grad_L_pi}
\nabla_\pi\ \mathcal{L}_{CVaR} & = &  \nabla_\pi\ H_\alpha(D^\pi(\xi|c(\mathcal{D})), \nu) \nonumber \\ 
                               & = &  \nabla_\pi\ \displaystyle\left[ \nu + \frac{1}{1-\alpha} \mathbb{E}_{\xi\sim\pi}\left[ (D^\pi(\xi|c(\mathcal{D}))-\nu)^{+}\right] \displaystyle\right] \nonumber \\
                               & = &  \frac{1}{1-\alpha}\nabla_\pi\ \mathbb{E}_{\xi\sim\pi}\left[ (D^\pi(\xi|c(\mathcal{D}))-\nu)^{+}\right] \nonumber \\
                               & = & \frac{1}{1-\alpha}\mathbb{E}_{\xi\sim\pi}\left[ (\nabla_\pi\ log P(\xi|\pi)) (D^\pi(\xi|c(\mathcal{D}))-\nu)^{+}\right]
\end{eqnarray}

\vspace{4mm}
\item \textbf{Gradient of $\mathcal{L}_{CVaR}$ w.r.t. $\nu$:}

\begin{eqnarray}
\label{eq:grad_L_nu}
\nabla_{\nu}\ \mathcal{L}_{CVaR} & = & \nabla_{\nu}\displaystyle\left[ \nu + \frac{1}{1-\alpha} \mathbb{E}_{\xi\sim\pi}\left[ (D^\pi(\xi|c(\mathcal{D}))-\nu)^{+}\right] \displaystyle\right] \nonumber \\
                                         & = &  1 + \frac{1}{1-\alpha} \mathbb{E}_{\xi\sim\pi}\left[\nabla_{\nu}\ (D^\pi(\xi|c(\mathcal{D}))-\nu)^{+}\right]  \nonumber \\
                                         & = & 1 - \frac{1}{1-\alpha} \mathbb{E}_{\xi\sim\pi}\left[ \mathbf{1}(D^\pi(\xi|c(\mathcal{D}))\geq\nu)\right]
\end{eqnarray}

\end{itemize}

\newpage

\subsection{Additional figures}
\label{appendix:analysis}

\begin{figure}[!h]
\centering
	\includegraphics[width=0.56\linewidth]{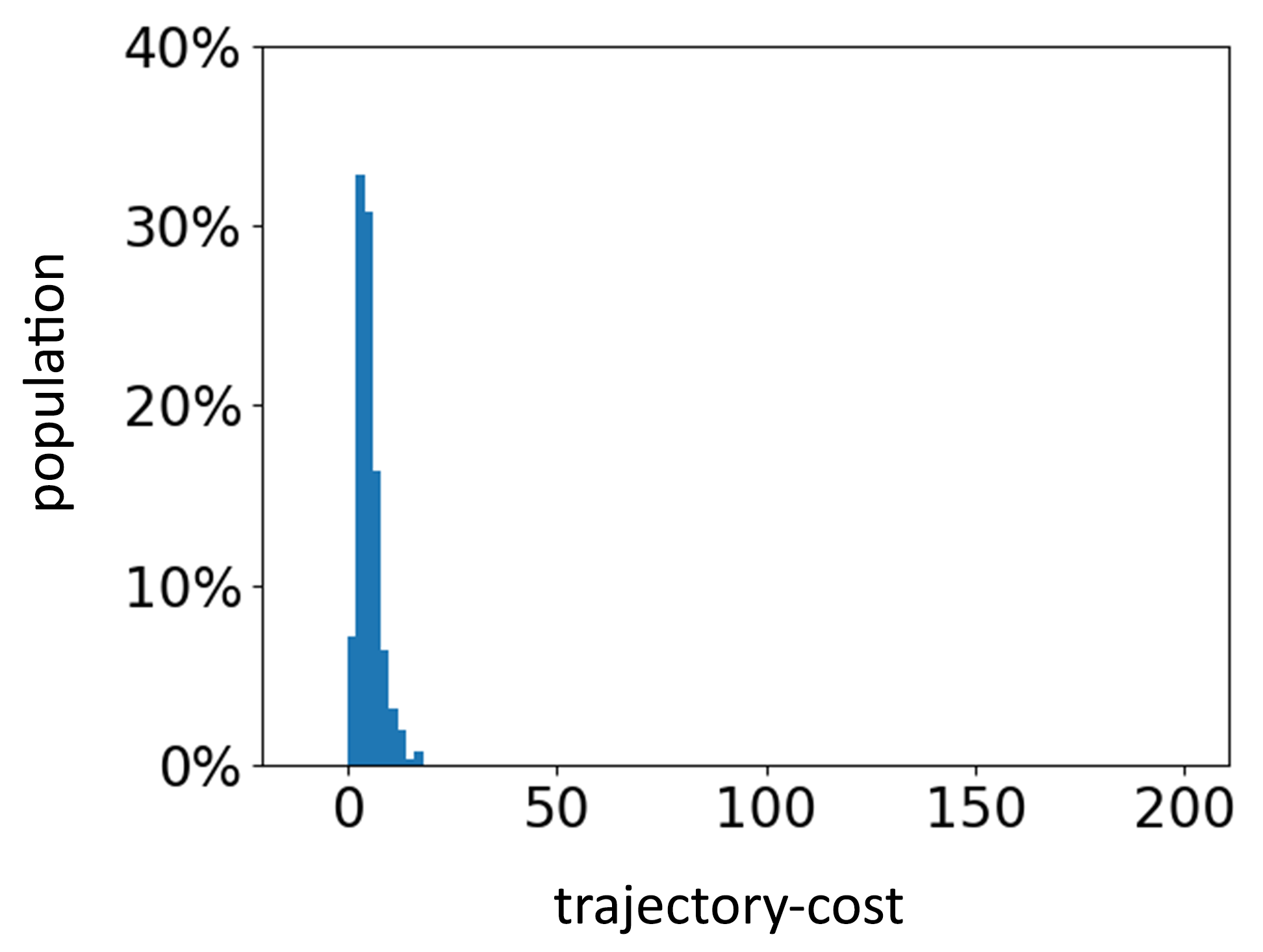}
	\caption{Histogram of costs of 250 trajectories generated by a GAIL-learned policy for Reacher-v1. The distribution shows no heavy tail. From Table~\ref{table:results} and Figure~\ref{fig:mean}, we observe that RAIL performs as well as GAIL even in cases where the distribution of trajectory costs is not heavy-tailed.}
    \label{fig:reacher}
\end{figure}

\end{document}